\documentclass [smallextended, natbib] {svjour3}
\usepackage[utf8]{inputenc}

\begin{document}
\title{A Comparative Analysis of Social Network Pages by Interests of Their Followers}
\titlerunning{Comparison of Interest Classifying Models}
\author{Elena Mikhalkova \and Nadezhda Ganzherli \and Yuri Karyakin}
\institute{E. Mikhalkova \at
              Tyumen State University, Tyumen, Russia \\
              \email{e.v.mikhalkova@utmn.ru}
           \and
           N. Ganzherli \at
              Tyumen State University, Tyumen, Russia \\
              \email{n.v.ganzherli@utmn.ru}  %CHANGE EMAIL
                         \and
           Yu. Karyakin \at
              Tyumen State University, Tyumen, Russia \\
              \email{y.e.karyakin@utmn.ru}
}

\date{Received: date / Accepted: date}

\maketitle

\begin{abstract}
Being a matter of cognition, user interests should be apt to classification independent of the language of users, social network
and content of interest itself.
To prove it, we analyze a collection of English and Russian Twitter and Vkontakte community pages by interests of their followers.
First, we create a model of Major Interests (MaIs) with the help of expert analysis and then classify a set of pages
using machine learning algorithms (SVM, Neural Network, Naive Bayes, and some other). We take three interest domains
that are typical of both English and Russian-speaking communities: football, rock music, vegetarianism. 
The results of classification show a greater correlation between Russian-Vkontakte and Russian-Twitter pages while English-Twitter
pages appear to provide the highest score.
\end{abstract}

\section{Introduction}

Social networks provide people with an opportunity to form social clusters that share interests not only sporadically 
but on a regular basis (circles of fans of different music, books, kinds of sports, etc.). Every circle communicates 
these interests creating lots of linguistic data to attract new followers and support interests of the existing ones.
Researchers often use these data in content-based user models to classify interests of particluar users. As a rule,
such models are tested on a corpus of one language downloaded from one social network. However, being a matter of cognition, 
user interests should be independent of the language in which they are expressed and the network where users communicate them,
when we try to process them with different algorithms. To see if the performance of machine learning algorithms is the same for 
two different languages and two networks, we will test them in three internationally popular interest 
domains: football, rock music, vegetarianism. For the present research, we collected three datasets from two different 
networks: the English (I) and Russian (II) corpora from Twitter and the Russian corpus (III) from Vkontakte, and built a linguistic 
model of user interests. Then, we tested the model with such machine learning instruments 
as SVM, Neural Network, Naive Bayes etc. to classify datasets according to the interests of users.

\section{Interest discovery by means of NLP}

In present, there exists a variety of content-based models of user interests. These models make use
of keywords, interests enlisted in profiles, tags attached to posts etc. Such data serve as the classification basis in 
works of \cite{bonhard2006knowing}, \cite{firan2007benefit}, \cite{dugan2007dogear}, \cite{li2008tag}, \cite{sen2009tagommenders}, 
\cite{guy2010social}, and many others.~\footnote{In recommender systems, tags and keywords in profiles define a scope of users that 
share similar interests. According to \cite{guy2009personalized}, this process is called {\em collaborative} filtering. 
\cite{pazzani1999framework} suggests {\em demographic} filtering that infers types of users with a common interest based on their 
age, gender, education etc. mentioned in profiles. With the rise of the social network analysis, many researchers, for example 
\cite{groh2007recommendations}, attempt to objectivize real-world communities and build recommender systems
with the help of social graphs ({\em social} filtering). A more detailed account of these approaches is given by \cite{burke2002hybrid}.} 
However, as these data are often very unreliable and hard to formalize,
classifying siocial network pages by interests of their users is not a trivial task.

Interest discovery has now become a separate branch of user modelling.
In regard to social networks, Natural Language Processing provides several approaches to interest discovery:
collecting interests as topics or terms from tweets, posts and messages; defining semantic relations between
keywords and searching for their correspondences in ontologies.

\cite{piao2011feasibility} view interests as terms and named entities extracted from a collection of user tweets.

In works of~\cite{mccallum2005topic},~\cite{ramage2010characterizing},~\cite{ahmed2011scalable}, 
interests are viewed as topics distributed across users' tweets.
The authors apply variations of Latent Dirichlet Allocation suggested by \cite{blei2003latent} 
as the main method of topic analysis to scale
user messages down to a particular topic. \cite{wang2014user} describe 
the User Message Model that is designed particularly for microblogs to reduce data sparseness
and topic diversity.

Interests can be represented as concepts in an ontology. The latter often includes named entities.
\cite{bakalov2009hybrid} suggest a hybrid user model that makes use of ontologies to specify
user interests. Interests are either extracted as keywords from the content of visited pages
or can be manually specified by a user. \cite{al2012interests} describe another approach
where the system creates a semantic graph of interests based on the ``entities'' mentioned in tweets. 
Entities are words denoting real-world phenomena that have an encyclopaedic description. For reference,
the authors used the currently deprecated knowledge base Freebase.~\footnote{http://www.freebase.com. 
Before the widespread use of knowledge bases, linguists often referred to WordNet, for example \cite{stefani1999exploiting}. 
More recent approaches like \cite{shen2013linking} use DBpedia.} A recent study of \cite{Piao2016} demonstrates that ``concept-based 
representations 
of user interests using a KB'' add efficiency to the model, but then there is no need to add ``rich semantic information from a 
KB to extend the interests of users.'' 

\section{Modelling social nature of interests}

It appears that interest discovery in social networks is a two-sided problem. First, regarding the number of published posts and comments,
although in social networks linguistic content is abundant, it is often very hard to structure.
Second, user interests themselves are an arcane matter: some researchers view them as topics, 
tags, keywords, etc. We will call the interest that attracts users to a page, the Major Interest (MaI). In the present research, we will 
attempt to classify a number of community pages based on three MaIs: football, rock music, vegetarianism. 

\subsection{Community pages}

In our research, we will focus on community pages, e.g. accounts of public value that represent institutions, authorities, famous people, 
leaders of social groups, events, etc. They exist in all networks known to us (Twitter, Vkontakte~\footnote{https://vk.com/. One of 
the most popular Russian social networks.}, Facebook, LiveJournal etc.). Many researchers already 
use data from such pages together with a user's individual page content but view them as complementary material.
Usually, but not necessarily, such accounts have many followers (typically, more than 1,000).

Concerning the content downloaded for analysis, from Vkontakte, we obtained posts, comments to posts, and comments from the so-called
``board''. As for Twitter, the only content available there is tweets.

\subsection{Data survey}

Observations show that for an expert it is quite easy to bind a community page to one certain MaI based on user comments and tweets and 
to find other pages with a similar MaI (the same kind of sports, music style, etc.). Many pages even provide links to other 
recommended pages. However, on the same page, users can mention a variety of different interest domains especially if they are related
hyponymically (a style of music and its substyles), antonymically (a football team vs. its opponent in a championship), 
pragmatically (a fooball team and a stadium where it trains). Therefore, to define the basis of classification, i.e. MaIs that are
not just microtopics and the pages that are devoted to these MaIs, we conducted an expert-based survey.

First, we downloaded comments from 20,000 random Vkontakte community pages. 4,460 pages contained texts of size
from 1 to 100,523 words. We cut down the number of pages to 4,000 leaving out pages with the smallest number of words.
So far, there was no automatic sorting of pages into spammed, flooded etc. Next, we asked a sociologist and a marketing
specialist to look through these pages and find several active communities with common interests, i.e. such community
pages where people actively interact about something they share an interest for. The result set included four communities
whose MaI is one of the following  1. rock music, 2. historical reenactment, 3. football, 4. vegetarianism. In addition, all these MaIs
are international and can be represented by pages in Russian as well as in English. We chose sample discussions from 
Vkontakte pages where people talk about things related to these MaIs. For control, a sample with several 
disparate objects of interest was chosen.

10 experts (linguists, sociologists, marketing specialists) gave their opinion on what community manifests itself in every
sample. We instructed experts to define if authors in the sample dialogue {\em are} a community and, if yes, 
explain why they think so. Thus, the expert answers were formulated freely without the aim of interest attribution. 
Some of them preferred to just name the community (``vegans'', ``rockers''); some stated the object of interest 
(``vegetarianism'', ``rock music''). If these keywords were mentioned, we assigned 1 point to the answer (a True Positive answer); if no or 
some other keywords were mentioned (``music addicts'' instead of ``rockers''), we assigned 0 points. The answers were 
put in a ranking table~\ref{tab:CV}. To see which samples relate to the most unanimous decision, we calculated percentage of
True Positive answers in every column (percent agreement).

\begin{table}
\label{tab:CV}
\caption{Percent agreement for expert analysis of community adherence}
\begin{center}
\begin{tabular}{lllllll}
\hline
Expert No. & Rock & Reenactment & Football & Vegetarianism & Control \\
\hline
1 & 1 & 1 & 1 & 1 & 0 \\
2 & 0 & 1 & 1 & 1 & 1 \\
3 & 1 & 1 & 1 & 1 & 1 \\
4 & 1 & 1 & 1 & 1 & 1 \\
5 & 0 & 1 & 1 & 1 & 1 \\
6 & 1 & 1 & 1 & 1 & 1 \\
7 & 1 & 1 & 1 & 1 & 1 \\
8 & 0 & 1 & 1 & 1 & 1 \\
9 & 0 & 1 & 1 & 1 & 1 \\
10 & 0 & 1 & 1 & 1 & 1 \\
\hline
Agreement, \% & 50 & \textbf{100} & \textbf{100} & \textbf{100} & 90 \\
\hline
\end{tabular}
\end{center}
\end{table}

Determining adherence of the authors of comments to communities of football fans, vegetarians, and historical reenactors, 
the raters showed perfect agreement. Fans of rock music were not as easy to define (50\% of raters recognized them). 
The control group also provided a highly reliable 
result~\footnote{We assigned 1 point for this sample if the expert directly expressed doubt in describing the community
or just wrote ``Don't know'' or left the field blank.} that allows us to state that the raters were not apt to see 
communities in any text we offer them.

\section{Community pages classification}

We used several machine learning algorithms to classify community pages that represent one of the mentioned MaIs.
As these interest domains are popular in the both English and Russian-speaking communities, we used Twitter to create the 
text collection in English and Russian, and Vkontakte, a very popular Russian social network, for the dataset in Russian.
However, we were unable to find any popular Twitter accounts devoted to historical reenactment (clubs, regiments, well-known reenactors)
in Russian. Consequently, we had to exclude this MaI from the further research. That leaves us with the three MaIs: 1. football, 
2. rock music, 3. vegetarianism.

For each class in the three corpora (I. English-Twitter, II. Russian-Twitter, III. Russian-Vkontakte), we prepared 30 texts downloaded 
from community pages. To normalize texts, we converted them to lowercase and removed punctuation marks, hashtags and emoji.~\footnote{We
could also transform every text into a set of microtopics or keywords, for example by using Latent Dirichlet Allocation, but, as the reader 
will see later, the current result was high enough without it.}

\subsection{Interclass classification}

To test performance of supervised machine learning algorithms on our collection, we randomly split the dataset into two equal sets of 15 
texts (training and test sets) so that these sets \textbf{do not overlap}. For vectors, we picked up the first 1,000 most frequent keywords 
including stop-words. We experimented with two vector models: Bernoulli (a simple absence or presence of a keyword in a text denoted by
0 or 1 correspondingly) and frequency distribution. We used plain frequencies of keywords in a text denoted by a whole number in the 
interval ${[0;+\infty)}$ and also normalized frequencies in the interval ${[0;1]}$. 

Classification algorithms that we chose for the survey are often met in NLP tasks like spam detection, sentiment analysis and the like:
Naive Bayes, Support Vector Machine, Neural Network etc. 
In particular, we used their implementation in the Python library Scikit-learn
described by \cite{scikit-learn}.~\footnote{Scikit-learn implementation of Neural Network for 
supervised learning is a type of Multi-layer Perceptron classifier with different optimization tools. As mentioned in the documentation,
``{\em lbfgs} is an optimizer in the family of quasi-Newton methods'' and {\em adam} is ``a stochastic gradient-based optimizer''.
For more references to particular methods, see Scikit-learn documentation at http://scikit-learn.org .} 
Table~\ref{tab:Interclass} demonstrates average results of F1-score in five tests. In every new test, train and test 
sets were randomly created anew.~\footnote{Some of the algorithms appeared to give
similar results regardless of what texts went to the train set. This is most obvious with the Multinomial Bayes that returned the same
result with the three models (Bernoulli and frequency). Results of some other algorithms varied more around the mean.} 

\begin{table}
\label{tab:Interclass}
\caption{Interclass classification: \={F1}-score. F - football, R - rock music, V - vegetarianism, T - Twitter,
Vk - Vkontakte, En - English, Ru - Russian, SVM - Support Vector Machine, lin. - linear kernel, pol. - polynomial kernel,
rad. - Radial Basis Function kernel, sig. - sigmoid kernel, Neur. - Neural Network, NB - Naive Bayes, Bern. - Bernoulli,
Mult. - Multinomial, Gaus. - Gaussian, LR - Logistic Regression, DT - Decision Trees, K-N - K-Neighbours. $\uparrow$ marks 
cases where normalization was more effective compared to plain frequencies.}
\begin{center}

\begin{tabular}{c}
\textbf{Bernoulli model}
%\hline
\end{tabular}

\begin{tabular}{l|lll|lll|lll}
\hline
 & & Vk Ru & & & T Ru & & & T En & \\
\hline
 & F & R & V & F & R & V & F & R & V \\
\hline
SVM lin.         & 0.958 & 0.974 & 0.968 & 1.0 & 0.98 & 0.982 & 1.0 & 0.988 & 0.988 \\
SVM pol.         & 0.822 & 0.854 & 0.898 & 0.918 & 0.898 & 0.86 & 0.994 & 0.988 & 0.994 \\
SVM rad.         & 0.952 & 0.954 & 0.994 & 1.0 & 0.982 & 0.982 & 1.0 & 0.994 & 0.994 \\
SVM sig.         & 0.952 & 0.954 & 0.994 & 1.0 & 0.988 & 0.988 & 1.0 & 0.994 & 0.994 \\
Neur. lbfgs      & 0.986 & 0.994 & 0.994 & 1.0 & 1.0 & 1.0 & 1.0 & 0.988 & 0.988 \\
Neur. adam       & 0.988 & 0.994 & 0.994 & 0.994 & 0.976 & 0.968 & 1.0 & 0.988 & 0.988 \\
Bern. NB         & 0.988 & 1.0 & 0.988 & 1.0 & 0.914 & 0.886 & 1.0 & 0.988 & 0.988 \\
Mult. NB         & 0.982 & 0.98 & 1.0 & 1.0 & 1.0 & 1.0 & 1.0 & 0.988 & 0.988 \\
Gaus. NB         & 0.968 & 0.988 & 0.982 & 1.0 & 0.982 & 0.978 & 1.0 & 0.982 & 0.982 \\
LR               & 1.0 & 1.0 & 1.0 & 1.0 & 1.0 & 1.0 & 1.0 & 0.988 & 0.988 \\
DT               & 0.836 & 0.882 & 0.792 & 0.98 & 0.988 & 0.968 & 0.924 & 0.872 & 0.9 \\
K-N              & 0.84 & 0.904 & 0.864 & 0.53 & 0.41 & 0.652 & 1.0 & 0.982 & 0.982 \\

\hline
\end{tabular}

\begin{tabular}{c}
\textbf{Frequency model}
%\hline
\end{tabular}

\begin{tabular}{l|lll|lll|lll}
\hline
 & & Vk Ru & & & T Ru & & & T En & \\
\hline
 & F & R & V & F & R & V & F & R & V \\
\hline
SVM lin.         & 0.79 & 0.82 & 0.874 & 0.988 & 0.914 & 0.916 & 0.982 & 0.956 & 0.962 \\
SVM pol.         & 0.468 & 0.486 & 0.664 & 0.932 & 0.79 & 0.832 & 0.976 & 0.962 & 0.95 \\  %R
SVM rad.         & 0.532 & 0.654 & 0.876 & 0.644 & 0.454 & 0.558 & 0.754 & 0.582 & 0.27 \\ %R
SVM sig.         & 0.43 & 0.046 & 0.148 & 0.268 & 0.266 & 0.222 & 0.054 & 0.208 & 0.514 \\ %R
Neur. lbfgs      & 0.912 & 0.924 & 0.926 & 0.898 & 0.808 & 0.838 & 0.942 & 0.912 & 0.756 \\  %R
Neur. adam       & 0.794 & 0.702 & 0.942 & 0.97 & 0.908 & 0.836 & 0.968 & 0.882 & 0.876 \\  %R
Bern. NB         & 0.982 & 0.994 & 0.976 & 1.0 & 0.928 & 0.914 & 1.0 & 0.982 & 0.982 \\
Mult. NB         & 0.982 & 0.98 & 1.0 & 1.0 & 1.0 & 1.0 & 1.0 & 0.988 & 0.988 \\
Gaus. NB         & 0.954 & 0.95 & 0.954 & 1.0 & 0.958 & 0.952 & 0.982 & 0.844 & 0.866 \\
LR               & 0.916 & 0.948 & 0.94 & 0.946 & 0.866 & 0.85 & 0.976 & 0.982 & 0.982 \\
DT               & 0.824 & 0.876 & 0.798 & 0.918 & 0.986 & 0.922 & 0.958 & 0.904 & 0.942 \\
K-N              & 0.646 & 0.662 & 0.802 & 0.83 & 0.752 & 0.788 & 0.974 & 0.91 & 0.892 \\ 

\hline
\end{tabular}

\begin{tabular}{c}
\textbf{Normalized frequency}
%\hline
\end{tabular}

\begin{tabular}{l|lll|lll|lll}
\hline
 & & Vk Ru & & & T Ru & & & T En & \\
\hline
 & F & R & V & F & R & V & F & R & V \\
\hline
SVM lin.         & 0.608 & 0.752 & 0.78 & 0.948 & 0.838 & 0.778 & 0.982 & {\em 0.974} & 0.958 \\  %R
SVM pol.         & {\em 0.612} & {\em 0.75} & {\em 0.8} & 0.916 & {\em 0.848} & 0.78 & 0.974 & {\em 0.968} & {\em 0.958} \\
SVM rad.         & 0.344 & {\em 0.668} & 0.67 & {\em 0.924} & {\em 0.834} & {\em 0.744} & {\em 0.964} & {\em 0.93} & {\em 0.964} \\
SVM sig.         & 0.344 & {\em 0.668} & {\em 0.67} & {\em 0.924} & {\em 0.834} & {\em 0.744} & {\em 0.964} & {\em 0.93} & {\em 0.964} \\
Neur. lbfgs      & {\em 0.928} & {\em 0.95} & 0.922 & {\em 0.948} & {\em 0.91} & {\em 0.89} & {\em 0.962} & {\em 0.946} & {\em 0.988} \\
Neur. adam       & {\em 0.976} & {\em 0.988} & {\em 0.976} & {\em 1.0} & {\em 0.958} & {\em 0.952} & {\em 1.0} & {\em 0.982} & {\em 0.982} \\
Bern. NB         & 0.398 & 0.8 & 0.836 & 0.954 & 0.738 & 0.412 & 0.988 & 0.956 & 0.954 \\  %R
Mult. NB         & 0.982 & 0.98 & 1.0 & 1.0 & 1.0 & 1.0 & 1.0 & 0.988 & 0.988 \\
Gaus. NB         & 0.954 & 0.95 & 0.954 & 1.0 & 0.958 & 0.952 & 0.982 & 0.844 & 0.866 \\ 
LR               & 0.818 & 0.852 & 0.86 & {\em 0.948} & 0.86 & 0.812 & 0.962 & 0.954 & 0.958 \\  %R
DT               & {\em 0.832} & {\em 0.92} & {\em 0.802} & {\em 0.966} & 0.968 & {\em 0.964} & 0.948 & 0.836 & 0.878 \\  %R
K-N              & {\em 0.66} & {\em 0.794} & 0.718 & 0.82 & {\em 0.762} & 0.568 & 0.97 & 0.89 & 0.904 \\  %R
\hline
\end{tabular}

\end{center}
\end{table}

Table~\ref{tab:Interclass} shows that Bernoulli model is the most effective one by mode: it has 29 scores of 1.0 when the two other models
have only 8 such scores each, and by mean: 0.958 against 0.819 for plain and 0.872 for normalized frequencies. The best performing 
algorithm is Linear Regression with Bernoulli model. The sum of its \={F1}-scores equals 8.976. The second best score (8.95) belongs to the
Neural Network ({\em lbfgs}). Multinomial Naive Bayes has the third best score (8.938) which is the same in all the three models. Hence,
we can assume that in our research this classifier is most insensitive to the model type, although intitially it was designed for word 
frequencies. But for Multinomial NB, Bernoulli models take the first 8 places in the ranking table. All things considered, we believe 
that Bernoulli models are the best solution for the tested linguistic model of interest classification.

Concerning normalization, it appears to be necessary for such algorithms as SVM with RBF and sigmoid kernels. Without it, they show
the lowest results (their sums of \={F1}-scores are 5.324 and 2.156 correspondingly). However, even with normalization, their peformance 
remains 
low compared to the winning solutions. Some well-performing models like Neural Network slightly increase their result with normalization.
As for Naive Bayes, it either gives the same result (Gaussian and Multinomial) or derates it (Bernoulli NB). Effects of normalization 
on the rest of the algorithms are not so obvious. For example, Linear Regression with normalization underperforms slightly in most of the 
cases, but even without normalization it is far below the top-score of Bernoulli model.

\subsection{Statistical analysis}

We will now try to analyze differences in classification of the three datasets according to the MaI, the language of user comminucation and 
the network where the texts were posted. For the analysis we will use the \={F1}-scores from Table~\ref{tab:Interclass}. 
First, we will normalize Table~\ref{tab:Interclass} excluding classifiers that gave lower results in the either of the two frequency models.
That leaves us with SVM (linear), Bernoulli NB, Linear Regression, Decision Trees with plain frequencies, and SVM (polynomial, sigmoid, RBF),
and Neural Networks with normalized frequencies. Multinomial and Gaussian NB have the same result in the both models.

For every MaI, the total sum of \={F1}-scores and sum dependent on the language and network is shown in Table~\ref{tab:Sum_MaIs}.

\begin{table}
\label{tab:Sum_MaIs}
\caption{Sums of MaI scores: \={F1}-score. F - football, R - rock music, V - vegetarianism, T - Twitter,
Vk - Vkontakte, En - English, Ru - Russian. \={x} denotes that  the value is given as the mean of the scores due to differences in the size 
of arrays (e.g. there are 48 \={F1}-scores per a MaI in Russian - Russian-Twitter and Russian-Vkontakte, and only 24 in English).}
\begin{center}

\begin{tabular}{l|llllllll}
\hline
MaI & Total & Vk Ru & T Ru & T En & Vk, \={x} & T, \={x} & Ru, \={x} & En, \={x} \\
\hline
F & 67.04 & 20.57 & 22.816 & 23.654 & 0.857 & 0.968 & 0.904 & 0.986 \\
R & 66.7 & 21.732 & 21.906 & 23.062 & 0.906 & 0.937 & 0.909 & 0.961 \\
V & 66.81 & 21.85 & 21.716 & 23.244 & 0.910 & 0.937 & 0.908 & 0.966 \\
\hline
\end{tabular}

\end{center}
\end{table}

To analyze significance of differences in the total scores, we used Mann-Whitney test. The median values of the three sets are F=0.982, 
R=0.971, V=0.968, the size of each set is 72. Football has the best total score. Mann-Whitney ${U}$ for Rock and Vegetarian sets demonstrates 
that they are likely to come from the same distribution: statistic=2562.0, pvalue=0.904, two-sided. However, there are significant differences
in Rock-Football (statistic=3130.5, pvalue=0.03) and Vegetarianism-Football (statistic=3107.5, pvalue=0.038) scores. Hence, we can assume 
that Football as a MaI was more supple to classification. It was also well-classified by the experts in the experiment.

However, there is one case where Football got the lowest score: the Russian-Vkontakte set. Mann-Whitney ${U}$ for Russian-Vkontake and 
Russian-Twitter Football sets shows a significant difference in the two distributions (statistic=151.5, pvalue=0.004,
two-sided). Meanwhile, Russian-Twitter and English-Twitter Football sets are more similar (statistic=334.5, pvalue=0.3,
two-sided) as well as Russian-Vkontakte and Russian-Twitter Rock sets (statistic=269.0, pvalue=0.695, two-sided). This prompts 
us to conclude that the Russian-Vkontakte Football set is not so representative as other sets although the reasons of its defects
are unclear.

Concerning the correlation between the experiment and classification, in case of Rock music, it did provide the lowest total result
of percent agreement (50\%) as well as of \={F1}-score (66.7). However, in automatic classification the difference is not as significant as 
in experiment where it is twice lower than for the other two MaIs). There can be two explanations: either there was a flaw in the experiment 
(e.g. the text suggested for the expert 
analysis was not so representative in features, some experts were unfamiliar with rock music, etc.) or some MaIs (like Rock music) are less 
supple to classification. Nevertheless, as we showed previously, Mann-Whitney ${U}$ for 
Rock and Vegetarianism demonstrates similarity in their scores. Hence, the experiment settings are more likely to have caused the
discrepance.

As for the factors of language and network, it appears that the mean values for Twitter are greater than those for Vkontakte, and the mean
values for English are greater than those for Russian in case of all the MaIs. Lower Vkontakte results can be caused by more noise features 
like spam, URLs, flood-messages compared to Twitter, as we used different software to clean the texts in Twitter and Vkontakte. 
However, considered separately,
Russian-Vkontakte and Russian-Twitter sets are more similar to each other than to English-Twitter by mean (0.953 and 0.963 versus 0.982
correspondingly) and median (0.891 and 0.923 versus 
0.972), which leads us to the idea that it is the language that brings more noise to the analysis. We conducted a series of tests
with normalization of the Russian text (stemming and part-of-speech classification), but they decreased results.

\section{Conclusion}

Summing up, the assumption about classification neutrality has proved wrong. First of all, there are slight changes in interest 
classification across networks, which can be due to amounts of noise features coming from spam, flood, attached content, etc.
Second, there are greater differences bound to the language: as far as our language model of MaI is concerned, pages in Russian
are harder to classify than pages in English. As for the interests themselves, it is hard to draw any strict conclusion. Some of 
the MaIs we took for research show a strong correlation of classification results, some stand aside. This is also an issue in
expert classification. 

Concerning the algorithms of classification, we faced the efficiency of the Bernoulli model. I.e. word frequencies are
not as important in classification as the absence or presence of characteristic features. And the more unique features are present in 
a page, the better. Hence, Linear Regression also came out to be the most effective algorithm. As for the frequency model,
tests showed that with some classifiers it needs normalization, but even normalized frequencies are not so representative. 
Generally speaking, we tend to think that MaIs are more like umbrella terms to a variety of topics discussed by communities rather 
than a class term semantically bound to page subtopics. Therefore, common NLP techniques do not work with them as they do in other 
NLP tasks.

We also discovered that objects of interest even though they can be spread in different parts of the world and communicated in 
different languages do not necessarily appear in all popular networks. In case with the historical reenactors, Twitter abounds in their
accounts in English, but, as far as we know, does not have a single reenactment account in Russian. Russian vegetarian accounts are also
scarce in Twitter. Unsurprisingly, Vkontakte has no living accounts of English-speaking reenactors and vegetarians. However, there are 
some English-language football fans accounts. There are also some concerns in regard to the size of the dataset. Due to the mentioned 
peculiarities, pages with veracious content representing certain MaIs are not easy to collect 
(except, probably, football).

\bibliographystyle{spbasic}
\bibliography{Interest_classified}

\end{document}